\documentstyle[psfig]{article}
\author{Fernando C. N. Pereira \and Rebecca N. Wright \\ AT\&T Research \\
600 Mountain Ave., Murray Hill, NJ 07974}
\date{\today}
\title{Finite-State Approximation of Phrase-Structure Grammars}

\newcommand{\tuple}[1]{{\langle #1 \rangle}}
\newcommand{\unf}[2]{[#1]^{#2}}

\newtheorem{proposition}{Proposition}

\newcommand{\QED}{\hfill \mbox{$\Box$}}
\newenvironment{proof}{\noindent {\bf Proof:}}{\nopagebreak \QED}

\newcommand{\fsa}{{\cal F}}
\newcommand{\cm}{{\cal M}}
\newcommand{\rec}{{\cal R}}
\newcommand{\Ra}{\Rightarrow}
\newcommand{\derives}{\stackrel{\ast}{\Ra}}
\newcommand{\ra}{\rightarrow}
\newcommand{\const}[1]{\mbox{\rm{}#1}}

\begin{document}
\maketitle
\begin{abstract}
Phrase-structure grammars are effective models for important syntactic
and semantic aspects of natural languages, but can be computationally
too demanding for use as language models in real-time speech
recognition. Therefore, finite-state models are used instead, even
though they lack expressive power. To reconcile those two
alternatives, we designed an algorithm to compute finite-state
approximations of context-free grammars and context-free-equivalent
augmented phrase-structure grammars. The approximation is exact for
certain context-free grammars generating regular languages, including
all left-linear and right-linear context-free grammars. The algorithm
has been used to build finite-state language models for limited-domain
speech recognition tasks.
\end{abstract}

\section{Motivation}

Grammars for spoken language systems are subject to the conflicting
requirements of language modeling for recognition and of language
analysis for sentence interpretation. For efficiency reasons, most
current recognition systems rely on finite-state language models.
These models, however, are inadequate for language interpretation,
since they cannot express the relevant syntactic and semantic
regularities.  Augmented phrase structure grammar (APSG) formalisms,
such as unification grammars \cite{Shieber:intro}, can express many of
those regularities, but they are computationally less suitable for
language modeling because of the inherent cost of computing state
transitions in APSG parsers.

The above conflict can be alleviated by using separate grammars for
language modeling and language interpretation. Ideally, the
recognition grammar should not reject sentences acceptable by the
interpretation grammar and as far as possible it should enforce
the constraints built into the interpretation grammar. However, if the
two grammars are built independently, those goals are difficult to
maintain.  For that reason, we have developed a method for
approximating APSGs with finite-state acceptors (FSAs).  Since such an
approximation is intended to serve as language model for a
speech-recognition front-end to the real parser, we require it to be
{\em sound} in the sense that the approximation accepts all strings in
the language defined by the APSG. Without qualification, the term
``approximation'' will always mean here ``sound approximation.''

If no further requirements were placed on the closeness of the
approximation, the trivial algorithm that assigns to any APSG over the
alphabet $\Sigma$ the regular language $\Sigma^{\ast}$ would do, but
of course this language model is useless. One possible criterion for
``goodness'' of approximation arises from the observation that many
interesting phrase-structure grammars have substantial parts that
accept regular languages. That does not mean that grammar rules are in
the standard forms for defining regular languages (left-linear or
right-linear), because syntactic and semantic considerations often
require that strings in a regular set be assigned structural
descriptions not definable by left- or right-linear rules. An ideal
criterion would thus be that if a grammar generates a regular
language, the approximation algorithm yields an acceptor for that
regular language.  In other words, one would like the algorithm to be
{\em exact} for all APSGs yielding regular languages. However, we will
see later that no such general algorithm is possible, that is, any
approximation algorithm will be inexact for some APSGs yielding
regular languages. Nevertheless, we will show that our method is exact
for left-linear and right-linear grammars, and for certain useful
combinations thereof.

\section{The Approximation Method}

Our approximation method applies to any context-free grammar (CFG), or
any constraint-based grammar \cite{Shieber:intro,Carpenter:logic} that
can be fully expanded into a context-free
grammar.\footnote{Unification grammars not in this class must first be
weakened using techniques such as Shieber's restrictor
\cite{Shieber:restriction}.} The resulting FSA accepts all
the sentences accepted by the input grammar, and possibly some
non-sentences as well.

The implementation takes as input unification grammars of a restricted form
ensuring that each feature ranges over a finite set. Clearly, such
grammars can only generate context-free languages, since an equivalent
CFG can be obtained by instantiating features in rules in all possible
ways.

\subsection{The Basic Algorithm}
The heart of our approximation method is an algorithm to convert the
LR(0) {\em characteristic machine} $\cm(G)$
\cite{Aho+Ullman:principles,Backhouse:syntax} of a CFG $G$ into an FSA
for a superset of the language $L(G)$ defined by $G$.  The
characteristic machine for a CFG $G$ is an FSA for the {\em viable
prefixes} of $G$, which are just the possible stacks built by the
standard shift-reduce recognizer for $G$ when recognizing strings in
$L(G)$.

This is not the place to review the characteristic machine
construction in detail. However, to explain the approximation
algorithm we will need to recall the main aspects of the construction.
The states of $\cm(G)$ are sets of {\em dotted rules}
$A\ra \alpha\cdot\beta$ where $A\ra
\alpha\beta$ is some rule of $G$. $\cm(G)$ is the determinization
by the standard subset construction \cite{Aho+Ullman:principles} of the
FSA defined as follows:
\begin{itemize}
\item The initial state is the dotted rule $S'\ra \cdot S$
where $S$ is the start symbol of $G$ and $S'$ is a new auxiliary start
symbol.

\item The final state is $S'\ra S\cdot$.

\item The other states are all the possible dotted rules of $G$.

\item There is a transition labeled $X$, where $X$ is a terminal or
nonterminal symbol, from $A\ra\alpha\cdot X\beta$
to $A\ra\alpha X\cdot\beta$.

\item There is an $\epsilon$-transition from $A\ra\alpha\cdot
B\beta$ to $B\ra\cdot\gamma$, where $B$ is a nonterminal
symbol and $B\ra\gamma$ is a rule in $G$.
\end{itemize}

$\cm(G)$ can be seen as the finite state control for a
nondeterministic shift-reduce pushdown recognizer $\rec(G)$ for $G$.
A state transition labeled by a terminal symbol $x$ from state $s$ to
state $s'$ licenses a {\em shift} move, pushing onto the stack of the
recognizer the pair $\tuple{s,x}$.  Arrival at a state containing a
{\em completed dotted rule} $A\ra \alpha\cdot$ licenses a {\em
reduction} move. This pops from the stack $|\alpha|$ elements,
checking that the symbols in the pairs match the corresponding
elements of $\alpha$, takes the transition labeled by A from the state
$s$ in the last pair popped, and pushes $\tuple{s,A}$ onto the
stack. (Full definitions of those concepts are given in Section
\ref{formal}.)

The basic ingredient of our approximation algorithm is the {\em
flattening} of a shift-reduce recognizer for a grammar $G$ into an FSA
by eliminating the stack and turning reduce moves into
$\epsilon$-transitions. It will be seen below that flattening
$\rec(G)$ directly leads to poor approximations in many interesting
cases. Instead, $\cm(G)$ must first be {\em unfolded} into a
larger machine whose states carry information about the possible
shift-reduce stacks of $\rec(G)$.  The
quality of the approximation is crucially influenced by how much stack
information is encoded in the states of the unfolded machine: too
little leads to coarse approximations, while too much leads to
redundant automata needing very expensive optimization.

\begin{figure}
\centerline{\psfig{figure=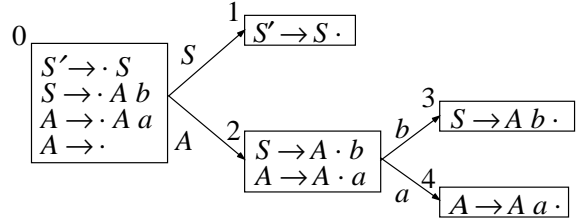,width=3in}}
\caption{Characteristic Machine for $G_1$}
\label{fig1}
\end{figure}

\begin{figure}
\centerline{\psfig{figure=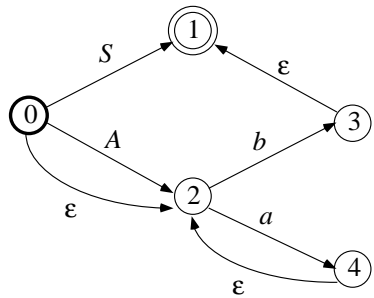,width=2.5in}}
\caption{Flattened Canonical Acceptor for $L(G_1)$}
\label{fig2}
\end{figure}

\begin{figure}
\centerline{\psfig{figure=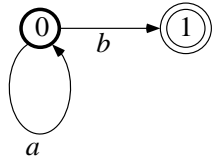,width=1.1in}}
\caption{Minimal Acceptor for $L(G_1)$}
\label{fig3}
\end{figure}

The algorithm is best understood with a simple example.
Consider
the left-linear grammar $G_1$
\[
\begin{array}{l}
S \ra Ab \\
A \ra A a \mid  \epsilon\qquad\mbox{.}
\end{array}
\]
$\cm(G_1)$ is
shown on Figure \ref{fig1}.
Unfolding is not required for this simple example, so the
approximating FSA is obtained from ${\cal M}(G_1)$ by the flattening
method outlined above. The reducing states in $\cm(G_1)$, those
containing completed dotted rules, are states 0, 3 and 4.  For
instance, the reduction at state 3 would lead to a $\rec(G_1)$ transition on
nonterminal $S$ to state 1, from the state that activated the rule
being reduced. Thus the corresponding $\epsilon$-transition goes from
state 3 to state 1.  Adding all the transitions that arise in this way
we obtain the FSA in Figure \ref{fig2}.
From this point on, the arcs labeled with nonterminals can be deleted,
and after simplification we obtain the deterministic finite
automaton (DFA) in Figure \ref{fig3},
which is the minimal DFA for $L(G_1)$.

\begin{figure}
\centerline{\psfig{figure=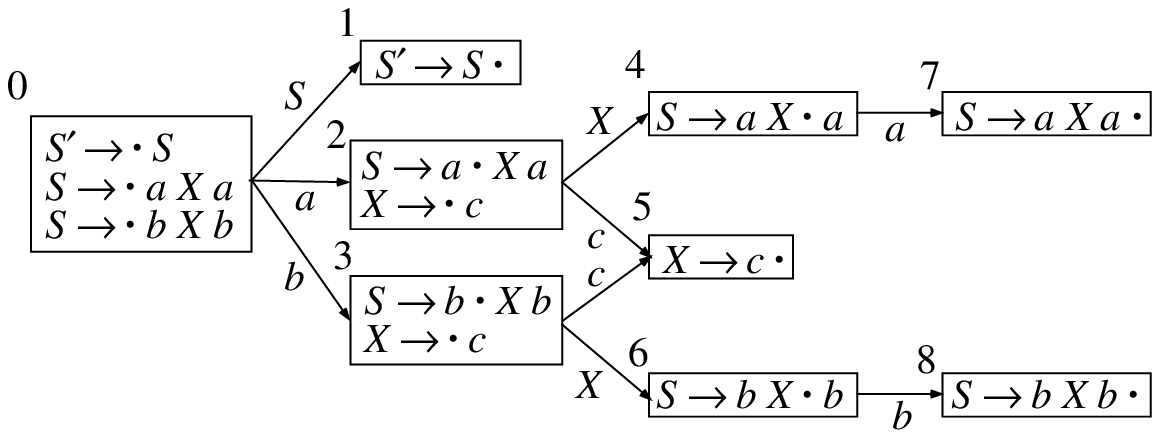,width=4in}}
\caption{Minimal Acceptor for $L(G_2)$}
\label{axa}
\end{figure}
\begin{figure}
\centerline{\psfig{figure=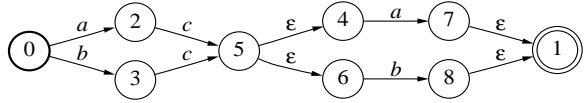,width=3in}}
\caption{Flattened Acceptor for $L(G_2)$}
\label{axa-flat}
\end{figure}
If flattening were always applied to the LR(0) characteristic machine
as in the example above, even simple grammars defining regular
languages might be inexactly approximated by the algorithm.  The
reason for this is that in general the reduction at a given reducing
state in the characteristic machine transfers to different states
depending on stack contents. In other words, the reducing state might
be reached by different routes which use the result of the reduction
in different ways. The following grammar $G_2$
\[
\begin{array}{l}
S \ra a X a \mid  b X b \\
X \ra c
\end{array}
\]
accepts just the two strings $aca$ and $bcb$, and has the characteristic
machine $\cm(G_2)$ shown in Figure \ref{axa}. However, the
corresponding flattened acceptor shown in Figure \ref{axa-flat} also accepts
$acb$ and $bca$, because
the $\epsilon$-transitions leaving state 5
do not distinguish between the different
ways of reaching that state encoded in the stack of $\rec(G_2)$.
\begin{figure}
\centerline{\psfig{figure=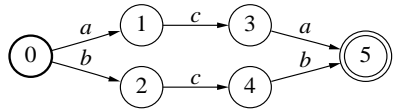,width=2in}}
\caption{Exact Acceptor for $L(G_2)$}
\label{axa-exact}
\end{figure}

Our solution for the problem just described is to unfold each state of the
characteristic machine into a set of states corresponding to different
stacks at that state, and flattening the corresponding recognizer
rather than the original one. Figure \ref{axa-exact} shows the
resulting acceptor for $L(G_2)$, now exact, after determinization and
minimization. 

In general the set of possible stacks at a state is
infinite. Therefore, it is necessary to do the unfolding not with
respect to stacks, but with respect to a finite partition of the set
of stacks possible at the state, induced by an appropriate equivalence
relation. The relation we use currently makes two stacks equivalent if
they can be made identical by {\em collapsing loops}, that is,
removing in a canonical way portions of stack pushed between two
arrivals at the same state in the finite-state control of the
shift-reduce recognizer, as described more formally and the end of
section \ref{soundness}. The purpose of collapsing a loop is to
``forget'' a stack segment that may be arbitrarily
repeated.\footnote{Since possible stacks can be shown to form a
regular language, loop collapsing has a direct connection to the
pumping lemma for regular languages.} Each equivalence class is
uniquely defined by the shortest stack in the class, and the classes
can be constructed without having to consider all the (infinitely)
many possible stacks. 

\subsection{Grammar Decomposition}
Finite-state approximations computed by the basic algorithm
may be extremely large, and their determinization, which is
required by minimization \cite{Aho+Hopcroft+Ullman:da},
can be computationally infeasible. These
problems can be alleviated by first decomposing the grammar to be
approximated into {\em subgrammars} and approximating the subgrammars
separately before combining the results. 

Each subgrammar in the decomposition of a grammar $G$ corresponds to a
set of nonterminals that are involved, directly or indirectly, in each
other's definition, together with their defining rules.  More
precisely, we define a directed graph $\const{conn}(G)$ whose nodes
are $G$'s nonterminal symbols, and which has an arc from $X$ to $Y$
whenever $Y$ appears in the right-hand side of one of $G$'s
rules and $X$ in the left-hand side. Each strongly connected
component of this graph
\cite{Aho+Hopcroft+Ullman:da} corresponds to a set of mutually
recursive nonterminals. 

Each nonterminal $X$ of $G$ is in exactly one strongly connected
component $\const{comp}(X)$ of $\const{conn}(G)$. Let
$\const{prod}(X)$ be the set of $G$ rules with left-hand sides
in $\const{comp}(X)$, and $\const{rhs}(X)$ be the set of right-hand
side nonterminals of $\const{comp}(X)$. Then the {\em defining
subgrammar} $\const{def}(X)$ of $X$ is the grammar with start symbol
$X$, nonterminal symbols $\const{comp}(X)$, terminal symbols $\Sigma
\cup (\const{rhs}(X)-\const{comp}(X))$ and rules
$\const{prod}(X)$. In other words, the nonterminal symbols not in
$\const{comp}(X)$ are treated as {\em pseudoterminal} symbols in
$\const{def}(X)$.

Each grammar $\const{def}(X)$ can be approximated with our basic
algorithm, yielding an FSA $\const{aut}(X)$. To see how to merge
together each of these subgrammar approximations to yield an
approximation for the whole of $G$, we observe first that the notion
of strongly connected component allows us to take each
$\const{aut}(X)$ as a node in a directed acyclic graph with an arc
from $\const{aut}(X)$ to $\const{aut}(X')$ whenever $X'$ is a
pseudoterminal of $\const{def}(X)$. We can then replace each
occurrence of a pseudoterminal $X'$ by its definition. More precisely,
each transition labeled by a pseudoterminal $X'$ from some state $s$
to state $s'$ in $\const{aut}(X)$ is replaced by
$\epsilon$-transitions from $s$ to the initial state of a separate
copy of $\const{aut}(X')$ and $\epsilon$-transitions from the final
states of the copy of $\const{aut}(X')$ to $s'$.  This process is then
recursively applied to each of the newly created instances of
$\const{aut}(X')$ for each pseudoterminal in $\const{def}(X)$. Since
the subautomata dependency graph is acyclic, the replacement process
must terminate.

\section{Formal Properties}
\label{formal}
We will show now that the basic approximation
algorithm described informally in the previous section is sound for
arbitrary CFGs and is exact for left-linear and right-linear CFGs.
From those results, it will be easy to see that the extended algorithm
based on decomposing the input grammar into strongly connected
components is also sound, and is exact for CFGs in which every
strongly connected component is either left linear or right linear.

In what follows, $G$ is a fixed CFG with terminal vocabulary $\Sigma$,
nonterminal vocabulary $N$ and start symbol $S$, and $V=\Sigma\cup N$.

\subsection{Soundness}
\label{soundness}
Let $\cm$ be the characteristic machine for $G$, with state set $ Q $,
start state $s_0$, final states $F$, and transition function
$\delta:S\times V\ra S$. As usual, transition functions such as
$\delta$ are extended from input symbols to input strings by defining
$\delta(s,\epsilon)=s$ and $\delta(s,\alpha\beta) =
\delta(\delta(s,\alpha),\beta)$.  The shift-reduce recognizer $\rec $
associated to $\cm $ has the same states, start state and final states
as ${\cal M}$. Its {\em configurations} are triples $\tuple{
s,\sigma,w}$ of a state, a stack and an input string. The stack is a
sequence of pairs $\tuple{s,X}$ of a state and a symbol. The
transitions of the shift-reduce recognizer are given as follows:
\begin{description}
\item[Shift:] $\tuple{ s,\sigma,xw}\vdash\tuple{
s',\sigma\tuple{s,x},w}$ if $\delta(s,x)=s'$

\item[Reduce:] $\tuple{ s,\sigma\tau,w}\vdash\tuple{
\delta(s',A),\sigma\tuple{s',A},w}$ if either
(1) $A\ra\cdot$ is a completed dotted rule in $s$, $s'=s$ and
$\tau$ is empty, or (2) $A\ra X_1\ldots X_n\cdot$ is a
completed dotted rule in $s$, $\tau=\tuple{s_1,X_1}\cdots\tuple{s_n,X_n}$ and
$s'=s_1$.
\end{description}
The {\em initial} configurations of $\rec $ are $\tuple{
s_0,\epsilon,w}$ for some input string $w$, and the {\em final}
configurations are $\tuple{ s,\tuple{s_0,S},\epsilon}$ for some state
$s\in F$. A {\em derivation} of a string $w$ is a sequence of
configurations $c_0,\ldots,c_m$ such that $c_0=\tuple{
s_0,\epsilon,w}$, $c_m$ is final, and $c_{i-1}\vdash c_i$ for $1 \le
i\le n$.

Let $s$ be a state. We define the set $\mbox{Stacks}(s)$ to contain
every sequence $\tuple{q_0,X_0}\ldots\tuple{q_k,X_k}$ such that $q_0 =
s_0$ and $q_i=\delta(q_{i-1},X_{i-1}), 1 \le i \le k$ and
$s=\delta(q_k,X_k)$.  In addition, $\mbox{Stacks}(s_0)$ contains the
empty sequence $\epsilon$. By construction, it is clear that if
$\tuple{ s,\sigma,w}$ is reachable from an initial configuration in
$\rec $, then $\sigma\in\mbox{Stacks}(s)$.

A {\em stack congruence} on $\rec $ is a family of equivalence
relations $\equiv_s$ on $\mbox{Stacks}(s)$ for each state $s\in{\cal
S}$ such that if $\sigma\equiv_s\sigma'$ and $\delta(s,X)=s'$ then
$\sigma\tuple{s,X}\equiv_{s'}\sigma\tuple{s,X}$. A stack congruence
$\equiv$ partitions each set $\mbox{Stacks}(s)$ into equivalence classes
$[\sigma]_s$ of the stacks in $\mbox{Stacks}(s)$ equivalent to
$\sigma$ under $\equiv_s$.

Each stack congruence $\equiv$ on $\rec $ induces a corresponding {\em
unfolded recognizer} $\rec_{\equiv} $. The states of the unfolded
recognizer are pairs $\tuple{s,[\sigma]_s}$, notated more concisely as
$\unf{\sigma}{s}$, of a state and stack equivalence class at that
state. The initial state is $\unf{\epsilon}{s_0}$, and the final
states are all $\unf{\sigma}{s}$ with $s\in F$ and
$\sigma\in\mbox{Stacks}(s)$. The transition function $\delta_{\equiv}$
of the unfolded recognizer is defined by
\[\delta_{\equiv}(\unf{\sigma}{s},X)=\unf{\sigma\tuple{s,X}}{\delta(s,X)}\qquad\mbox{.}\]
That this is well-defined follows immediately from the definition of
stack congruence.

The definitions of dotted rules in states, configurations, shift and
reduce transitions given above carry over immediately to unfolded
recognizers. Also, the characteristic recognizer can also be seen as
an unfolded recognizer for the trivial coarsest congruence.

Unfolding a characteristic recognizer does not change the language
accepted: 
\begin{proposition}\label{unfold}
Let $G$ be a CFG, $\rec$ its characteristic recognizer with transition
function $\delta$, and $\equiv$ 
a stack congruence on $\rec$. Then
$\rec_{\equiv}$ and $\rec$ are equivalent.
\end{proposition}
\begin{proof}
We show first that any string $w$ accepted by $\rec_\equiv$ is
accepted by $\rec$. Let $d$ be
configuration of $\rec_\equiv$. By construction,
$d=\tuple{\unf{\rho}{s},\sigma,u}$,  with $\sigma =
\tuple{\tuple{q_0,e_0},X_0}\cdots \tuple{\tuple{q_k,e_k},X_k}$ for
appropriate stack equivalence classes $e_i$. We define
$\hat{d}=\tuple{s, \hat{\sigma}, u}$, with $\hat{\sigma}=
\tuple{q_0,X_0}\cdots \tuple{q_k,X_k}$. If $d_0,\ldots,d_m$ is a
derivation of $w$ in $\rec_\equiv$, it is easy to verify that
$\hat{d}_0,\ldots,\hat{d}_m$ is a derivation of $w$ in $\rec$.

Conversely, let $w\in L(G)$, and let $c_0,\ldots,c_m$ be a derivation of
$w$ in $\rec$, with $c_i=\tuple{s_i,\sigma_i,u_i}$. We define
$\bar{c}_i=\tuple{\unf{\sigma_i}{s_i},\bar{\sigma}_i,u_i}$, where
$\bar{\epsilon} = \epsilon$ and $\overline{\sigma\tuple{s,X}} =
\bar{\sigma}\tuple{\unf{\sigma}{s},X}$.

If $c_{i-1}\vdash c_i$ is a shift move, then $u_{i-1}=xu_i$ and
$\delta(s_{i-1},x) = s_i$. Therefore,
\begin{eqnarray*}
\delta_{\equiv}(\unf{\sigma_{i-1}}{s_{i-1}},x) & = &
\unf{\sigma_{i-1}\tuple{s_{i-1},x}}{\delta(s_{i-1},x)} \\
& = & \unf{\sigma_i}{s_i} \qquad\mbox{.}
\end{eqnarray*}
Furthermore,
\[
\bar{\sigma}_i = \overline{\sigma_{i-1}\tuple{s_{i-1},x}} =
\bar{\sigma}_{i-1}\tuple{\unf{\sigma_{i-1}}{s_{i-1}},x}\qquad\mbox{.}
\]
Thus we have
\begin{eqnarray*}
\bar{c}_{i-1} & =
&\tuple{\unf{\sigma_{i-1}}{s_{i-1}},\bar{\sigma}_{i-1},xu_i} \\
\bar{c}_i & = &
\tuple{\unf{\sigma_i}{s_i},\bar{\sigma}_{i-1}\tuple{\unf{\sigma_{i-1}}{s_{i-1}},x},u_i}
\end{eqnarray*}
with
$\delta_{\equiv}(\unf{\sigma_{i-1}}{s_{i-1}},x)=\unf{\sigma_i}{s_i}$.
Thus, by definition of shift move, $\bar{c}_{i-1}\vdash\bar{c}_i$ in
$\rec_{\equiv}$.

Assume now that $c_{i-1}\vdash c_i$ is a reduce move in $\rec$. Then
$ u_i = u_{i-1}$ and we have a state $s$ in $\rec$, a symbol $A\in
N$, a stack $\sigma$ and a sequence $\tau$ of state-symbol pairs such that
\begin{eqnarray*}
s_i & = & \delta(s,A) \\
\sigma_{i-1} & = &\sigma\tau \\
\sigma_i& = &\sigma\tuple{s,A}
\end{eqnarray*}
and either
\begin{itemize}
\item[(a)] $A\ra\cdot$ is in $s_{i-1}$, $s=s_{i-1}$ and
$\tau=\epsilon$, or
\item[(b)] $A\ra X_1\cdots X_n\cdot$ is in $s_{i-1}$	,
$\tau=\tuple{q_1,X_1}\cdots\tuple{q_n,X_n}$ and $s=q_1$.
\end{itemize}

Let $\bar{s} =\unf{\sigma}{s}$. Then
\begin{eqnarray*}
\delta_{\equiv}(\bar{s},A) & = & \unf{\sigma\tuple{s,A}}{\delta(s,A)}
\\
& = & \unf{\sigma_i}{s_i}
\end{eqnarray*}

We now define a pair sequence $\bar{\tau}$ to play the same role
in $\rec_\equiv$ as $\tau$ does in $\rec$. In case (a) above,
$\bar{\tau}=\epsilon$. Otherwise, let $\tau_1 = \epsilon$ and
$\tau_i=\tau_{i-1}\tuple{q_{i-1},X_{i-1}}$ for $2 \le i \le n$, and define
$\bar{\tau}$ by
\[
\bar{\tau}=\tuple{\unf{\sigma}{q_1},X_1}\cdots\tuple{\unf{\sigma\tau_i}{q_i},X_i}\cdots\tuple{\unf{\sigma\tau_n}{q_n},X_n}
\]
Then
\begin{eqnarray*}
\bar{\sigma}_{i-1}& = &\overline{\sigma\tau} \\
& = &
\overline{\sigma\tuple{q_1,X_1}\cdots\tuple{q_{n-1},X_{n-1}}}\tuple{\unf{\sigma\tau_n}{q_n},X_n}
\\
& = &
\overline{\sigma\tuple{q_1,X_1}\cdots\tuple{q_{i-1},X_{i-1}}}\tuple{\unf{\sigma\tau_i}{q_i},X_i}\cdots\tuple{\unf{\sigma\tau_n}{q_n},X_n} \\
& = &\bar{\sigma}\bar{\tau} \\
\bar{\sigma}_i & = & \overline{\sigma\tuple{s,A}} \\
& = & \bar{\sigma}\tuple{\unf{\sigma}{s},A} \\
& = & \bar{\sigma}\tuple{\bar{s},A}\qquad\mbox{.}
\end{eqnarray*}
Thus
\begin{eqnarray*}
\bar{c}_i & = &
\tuple{\delta_\equiv(\bar{s},A),\bar{\sigma}\tuple{\bar{s},A},u_i} \\
\bar{c}_{i-1} & = &
\tuple{\unf{\sigma_{i-1}}{s_{i-1}},\bar{\sigma}\bar{\tau},u_{i-1}}
\end{eqnarray*}
which by construction of $\bar{\tau}$ immediately entails that
$\bar{c}_{i-1}\vdash \bar{c}_i$ is a reduce move in $\rec_{\equiv}$.
\end{proof}

For any unfolded state $p$, let $\mbox{Pop}(p)$ be the set of states
reachable from $p$ by a reduce transition. More precisely,
$\mbox{Pop}(p)$ contains any state $p'$ such that there is a
completed dotted rule $A\ra \alpha\cdot$ in $p$ and a state
$p''$ containing $A\ra\cdot\alpha$ such that $\delta_{\equiv}(p'',\alpha)=p$ and
$\delta_{\equiv}(p'',A)=p'$.  Then the {\em flattening} ${\cal
F}_{\equiv} $ of $\rec_{\equiv} $ is an NFA
with the same state set, start state and final states as ${\cal
R}_{\equiv} $ and nondeterministic transition function $\phi_{\equiv}$ defined as
follows:
\begin{itemize}
\item If $\delta_{\equiv}(p,x)=p'$ for some $x\in\Sigma$, then
$p'\in\phi_{\equiv}(p,x)$
\item If $p'\in\mbox{Pop}(p)$ then $p'\in\phi_{\equiv}(p,\epsilon)$.
\end{itemize}

Let $c_0,\ldots,c_m$ be a derivation of string $w$ in $\rec $,
and put $c_i=\tuple{ q_i,\sigma_i,w_i}$, and $p_i =\unf{\sigma_i}{p_i}$. By construction, if
$c_{i-1} \vdash c_i$ is a shift move on $x$ ($w_{i-1}=xw_i$), then
$\delta_{\equiv}(p_{i-1},x)=p_i$,
and thus
$p_i\in\phi_{\equiv}(p_{i-1},x)$.
Alternatively, assume the transition is a reduce move associated to
the completed dotted rule
$A\ra\alpha\cdot$. We consider first the case $\alpha\not=\epsilon$.
Put $\alpha=X_1\ldots X_n$. By definition of reduce move, there is a
sequence of states $r_1,\ldots,r_n$ and a stack $\sigma$ such that
$\sigma_{i-1} = \sigma\tuple{r_1,X_1}\ldots\tuple{r_n,X_n}$, $\sigma_i =
\sigma\tuple{r_1,A}$, $r_1$ contains $A\ra\cdot\alpha$, $\delta(r_1,A) = q_i$, and $\delta(r_j,X_j) = r_{j+1}$
for $1 \le j < n$. By definition of stack congruence, we will then
have \[\delta_{\equiv}(\unf{\sigma\tau_j}{r_j},X_j) =
\unf{\sigma\tau_{j+1}}{r_{j+1}} \]
where $\tau_1=\epsilon$ and $\tau_j =
\tuple{r_1,X_1}\ldots\tuple{r_{j-1},X_{j-1}} $ for $j > 1$.
Furthermore, again by definition of stack congruence we have
$\delta_{\equiv}(\unf{\sigma}{r_1},A) = p_i$. Therefore,
$p_i\in\mbox{Pop}(p_{i-1})$ and thus
$p_i\in\phi_{\equiv}(p_{i-1},\epsilon)$. A similar but simpler
argument allows us to reach the same conclusion for the case
$\alpha=\epsilon$. Finally, the definition of final state for ${\cal
R}_{\equiv} $ and ${\cal F}_{\equiv} $ makes $p_m$ a final state.
Therefore the sequence $p_0,\ldots,p_m$ is an accepting path for $w$
in ${\cal F}_{\equiv}$. We have thus proved

\begin{proposition}\label{sound}
For any CFG $G$ and stack congruence $\equiv$ on the canonical
{\rm LR(0)} shift-reduce recognizer $\rec(G)$ of $G$, $L(G)\subseteq
L({\cal F}_{\equiv}(G))$, where ${\cal F}_{\equiv}(G)$ is the
flattening of $\rec(G)_\equiv$.
\end{proposition}

To complete the proof of soundness for the basic algorithm, we must
show that the stack collapsing equivalence described informally
earlier is indeed a stack congruence. A stack $\tau$ is a {\em loop}
if $\tau=\tuple{s_1,X_1}\ldots\tuple{s_k,X_k}$ and
$\delta(s_k,X_k)=s_1$. A stack $\tau$ is a {\em minimal} loop if no
prefix of $\tau$ is a loop. A stack that contains a loop is {\em
collapsible}. A collapsible stack $\sigma$ {\em immediately collapses}
to a stack $\sigma'$ if $\sigma=\rho\tau\upsilon$,
$\sigma'=\rho\upsilon$, $\tau$ is a minimal loop and there is no other
decomposition $\sigma=\rho'\tau'\upsilon'$ such that $\rho'$ is a
proper prefix of $\rho$ and $\tau'$ is a loop. By these definitions, a
collapsible stack $\sigma$ immediately collapses to a unique stack
$C(\sigma)$. A stack $\sigma$ {\em collapses} to $\sigma'$ if
$\sigma'=C^n(\sigma)$.  Two stacks are equivalent if they can be
collapsed to the same uncollapsible stack. This equivalence relation
is closed under suffixing, therefore it is a stack congruence. Each
equivalence class has a canonical representative, the unique
uncollapsible stack in it, and clearly there are finitely many
uncollapsible stacks.

We compute the possible uncollapsible stacks associated with states as
follows. To start with, the empty stack is associated with the initial
state. Inductively, if stack $\sigma$ has been associated with state
$s$ and $\delta(s,X)=s'$, we associate $\sigma'=\sigma\tuple{s,X}$
with $s'$ unless $\sigma'$ is already associated with $s'$ or $s'$
occurs in $\sigma$, in which case a suffix of $\sigma'$ would be a
loop and $\sigma'$ thus collapsible. Since there are finitely many
uncollapsible stacks, the above computation is must terminate.

When the grammar $G$ is first decomposed into strongly connected
components $\const{def}(X)$, each approximated by $\const{aut}(X)$,
the soundness of the overall construction follows easily by induction
on the partial order of strongly connected components and by the
soundness of the approximation of $\const{def}(X)$ by
$\const{aut}(X)$, which guarantees that each $G$ sentential form over
$\Sigma \cup (\const{rhs}(X)-\const{comp}(X) $ accepted by
$\const{def}(X)$ is accepted by $\const{aut}(X)$.

\subsection{Exactness} \label{exactness}

While it is difficult to decide what should be meant by a ``good''
approximation, we observed earlier that a desirable feature of an
approximation algorithm is that it be exact for a wide class of
CFGs generating regular languages.  We show in this section that our
algorithm is exact for both left-linear and right-linear CFGs, and
as a consequence for CFGs that can be decomposed into independent left
and right linear components. On the other hand, a theorem of Ullian's
\cite{Ullian:partial-cfls} shows that there can be no partial
algorithm mapping CFGs to FSAs that terminates on every CFG yielding a
regular language $L$ with an FSA accepting exactly $L$.

The proofs that follow rely on the following basic definitions and
facts about the LR(0) construction. Each LR(0) state $s$ is the {\em
closure} of a set of a certain set of dotted rules, its {\em core}.
The closure $[R]$ of a set $R$ of dotted rules is the smallest set of
dotted rules containing $R$ that contains $B\ra\cdot\gamma$ whenever
it contains $A\ra\alpha\cdot B\beta$ and $B\ra\gamma$ is in $G$.  The
core of the initial state $s_0$ contains just the dotted rule $S'\ra
\cdot S$. For any other state $s$, there is a state $s'$ and a symbol
$X$ such that $s$ is the closure of the set core consisting of all
dotted rules $A\ra\alpha X\cdot\beta$ where $A\ra\alpha\cdot X\beta$
belongs to $s'$.

\subsubsection{Left-Linear Grammars}
A CFG $G$ is left-linear if
each rule in $G$ is of the form $A \ra B \beta$ or $A
\ra \beta$, where $A, B \in N$ and $\beta \in \Sigma^\ast$.

\begin{proposition}
Let $G$ be a left-linear CFG, and let $\fsa$ be the FSA derived from
$G$ by the basic approximation algorithm.  Then $L(G) = L(\fsa)$.
\end{proposition}

\begin{proof}
By Proposition \ref{sound}, $L(G) \subseteq L(\fsa)$.  Thus we
need only show $L(\fsa) \subseteq L(G)$.  

Since $\cm(G)$ is deterministic, for each $\alpha\in V^\ast$ there is
at most one state $s$ in $\cm(G)$ reachable from $s_0$ by a path
labeled with $\alpha$. If $s$ exists, we define
$\bar{\alpha}=s$. Conversely, each state $s$ can be identified with a
string $\hat{s}\in V^{\ast}$ such that every dotted rule in $s$ is of
the form $A\ra\hat{s}\cdot\alpha$ for some $A\in N$ and $\alpha\in
V^\ast$. Clearly, this is true for $s_0 = [S'\ra\cdot S]$, with
$\hat{s}_0=\epsilon$. The core $\dot{s}$ of any other state $s$ will
by construction contain only dotted rules of the form
$A\ra\alpha\cdot\beta$ with $\alpha\ne\epsilon$. Since $G$ is left
linear, $\beta$ must be a terminal string, thus $s=\dot{s}$. Therefore
every dotted rule $A\ra\alpha\cdot\beta$ in $s$ results from dotted
rule $A\ra\cdot\alpha\beta$ in $s_0$ by a unique transition path
labeled by $\alpha$ (since $\cm(G)$ is deterministic). This means that
if $A\ra\alpha\cdot\beta$ and $A'\ra\alpha'\cdot\beta'$ are in $s$, it
must be the case that $\alpha=\alpha'$.

To go from the characteristic machine $\cm(G)$ to the FSA $\fsa$, the
algorithm first unfolds $\cm(G)$ using the stack congruence relation,
and then flattens the unfolded machine by replacing reduce moves with
$\epsilon$-transitions.  However, the above argument shows that the
only stack possible at a state $s$ is the one corresponding to the
transitions given by $\hat{s}$, and thus there is a single stack
congruence state at each state. 
Therefore, $\cm(G)$ will only be flattened, not unfolded.  
Hence the transition function $\phi$ for the resulting flattened
automaton $\fsa$ is defined as follows, where
$\alpha \in N\Sigma^\ast \cup \Sigma^\ast, a \in \Sigma$, and $A \in N$:
\begin{enumerate}
\item[(a)] $\phi(\bar{\alpha}, a) = \{\overline{\alpha a}\}$
\item[(b)] $\phi(\bar{\alpha}, \epsilon) = \{\bar{A} \mid A \ra \alpha\in G\}$
\end{enumerate}
The start state of $\fsa$ is $\bar{\epsilon}$. The only final state is
$\bar{S}$.

We will establish the connection between $\fsa$
derivations and $G$ derivations. We claim that if there is a path from
$\bar{\alpha}$ to $\bar{S}$ labeled by $w$ then either there is a
rule $A\ra\alpha$ such that $w=xy$ and $S
\derives Ay \Ra \alpha xy$, or $\alpha = S$ and $w = \epsilon$.
The claim is proved by induction on $|w|$.  

For the base case, suppose $|w| = 0$ and there is a path from
$\bar{\alpha}$ to $\bar{S}$ labeled by $w$.  Then $w = \epsilon,$ and
either $\alpha = S$, or there is a path of $\epsilon$-transitions from
$\bar{\alpha}$ to $\bar{S}$.  In the latter case, $S \derives A
\Ra \epsilon$ for some $A \in N$ and rule $A \ra
\epsilon$, and thus the claim holds.

Now, assume that the claim is true for all $|w| < k$, and suppose
there is a path from $\bar{\alpha}$ to $\bar{S}$ labeled $w'$, for
some $|w'| = k$.  Then $w' = aw$ for some terminal $a$ and $|w| < k$,
and there is a path from $\overline{\alpha a}$ to $\bar{S}$ labeled
by $w$.  By the induction hypothesis, $S \derives Ay \Ra \alpha ax'y$,
where $A \ra \alpha ax'$ is a rule and $x'y = w$ (since $\alpha a \neq
S$).  Letting $x = ax'$, we have the desired result.

If $w \in L(\fsa)$, then there is a path from $\bar{\epsilon}$ to
$\bar{S}$ labeled by $w$.  Thus, by the claim just proved, $S \derives
Ay \Ra xy$, where $A \ra x$ is a rule and $w = xy$ (since $\epsilon
\neq S$).  Therefore, $S \derives w$, so $w \in L(G)$, as desired.
\end{proof}

\subsubsection{Right-Linear Grammars}
A CFG $G$ is  right linear if each rule in $G$ is
of the form $A \ra \beta B$ or $A \ra \beta$, where
$A, B \in N$ and $\beta \in \Sigma^\ast$.  

\begin{proposition}
Let $G$ be a right-linear CFG and $\fsa$ be the FSA derived from $G$
by the basic approximation algorithm.  Then $L(G) =
L(\fsa)$.\label{right-linear}
\end{proposition}

\begin{proof}
As before, we need only show $L(\fsa) \subseteq L(G)$.  

Let $\rec$ be the shift-reduce recognizer for $G$.  The key fact to
notice is that, because $G$ is right-linear, no shift transition may
follow a reduce transition.  Therefore, no terminal transition in
$\fsa$ may follow an $\epsilon$-transition, and after any
$\epsilon$-transition, there is a sequence of $\epsilon$-transitions
leading to the final state $[S'
\ra S \cdot]$.  Hence $\fsa$ has the following kinds of states: the
start state, the final state, states with terminal transitions
entering and leaving them (we call these {\em reading} states), states
with $\epsilon$-transitions entering and leaving them ({\em prefinal}
states), and states with terminal transitions entering them and
$\epsilon$-transitions leaving them ({\em crossover} states).  Any
accepting path through $\fsa$ will consist of a sequence of a start
state, reading states, a crossover state, prefinal states, and a final
state.  The exception to this is a path accepting the empty string,
which has a start state, possibly some prefinal states, and a final
state.

The above argument also shows that unfolding does not change the set
of strings accepted by $\fsa$, because any reduction in $\rec_\equiv$
(or $\epsilon$-transition in $\fsa$), is guaranteed to be part of a
path of reductions ($\epsilon$-transitions) leading to a final state
of $\rec_\equiv$ ($\fsa$).

Suppose now that $w = w_1 \ldots w_n$ is accepted by $\fsa$.  Then
there is a path from the start state $s_0$ through reading states
$s_1, \ldots, s_{n-1}$, to crossover state $s_n$, followed by
$\epsilon$-transitions to the final state. We claim that if there
there is a path from $s_i$ to $s_n$ labeled $w_{i+1} \ldots w_n$,
then there is a dotted rule $A \ra x \cdot yB$ in $s_i$ such $B
\derives z$ and $yz = w_{i+1} \ldots w_n$, where $A \in N, B \in N \cup
\Sigma^*, y,z \in \Sigma^*$, and one of the following holds:
\begin {enumerate}
\item[(a)] $x$ is a nonempty suffix of $w_1 \ldots w_i$,
\item[(b)] $x = \epsilon$, $A'' \derives A$, $A' \ra x' \cdot A''$ is a
dotted rule in $s_i$, and $x'$ is a nonempty suffix of $w_1 \ldots
w_i$, or
\item[(c)] $x = \epsilon$, $s_i = s_0$, and $S \derives A$.
\end{enumerate}

We prove the claim by induction on $n-i$. For the base case, suppose
there is an empty path from $s_n$ to $s_n$.  Because $s_n$ is the
crossover state, there must be some dotted rule $A
\ra x \cdot$ in $s_n$.  Letting $y = z = B = \epsilon$, we get
that $A \ra x \cdot yB$ is a dotted rule of $s_n$ and $B = z$.  The
dotted rule $A \ra x \cdot yB$ must have either been added to $s_n$ by
closure or by shifts.  If it arose from a shift, $x$ must be a
nonempty suffix of $w_1 \ldots w_n$. If the dotted rule arose by
closure, $x = \epsilon$, and there is some dotted rule $A' \ra x'
\cdot A''$ such that $A'' \derives A$ and $x'$ is a nonempty suffix of
$w_1
\ldots w_n$.

Now suppose that the claim holds for paths from $s_i$ to $s_n$, and
look at a path labeled $w_i \ldots w_n$ from $s_{i-1}$ to $s_n$.  By
the induction hypothesis, $A \ra x \cdot yB$ is a dotted rule of
$s_i$, where $B \derives z$, $uz = w_{i+1} \ldots w_n$, and (since
$s_i \neq s_0$), either $x$ is a nonempty suffix of $w_1 \ldots w_i$
or $x = \epsilon, A' \ra x' \cdot A''$ is a dotted rule of $s_i$, $A''
\derives A$, and $x'$ is a nonempty suffix of $w_1 \ldots w_i$.

In the former case, when $x$ is a nonempty suffix of $w_1 \ldots w_i$,
then $x = w_j \ldots w_i$ for some $1 \leq j < i$.  Then $A
\ra w_j \ldots w_i \cdot yB$ is a dotted rule of $s_i$, and
thus $A \ra w_j \ldots w_{i-1} \cdot w_i yB$ is a dotted rule
of $s_{i-1}$.  If $j \leq i-1$, then $w_j \ldots w_{i-1}$ is a nonempty
suffix of $w_1 \ldots w_{i-1}$, and we are done.  Otherwise, $w_j
\ldots w_{i-1} = \epsilon$, and so $A \ra \cdot w_iyB$ is a
dotted rule of $s_{i-1}$.  Let $y'$ = $w_iy$.  Then $A \ra \cdot
y'B$ is a dotted rule of $s_{i-1}$, which must have been added by
closure.  Hence there are nonterminals $A'$ and $A''$ such that $A''
\derives A$ and $A' \ra x' \cdot A''$ is a dotted rule of
$s_{i-1}$, where $x'$ is a nonempty suffix of $w_1 \ldots w_{i-1}$.

In the latter case, there must be a dotted rule $A' \ra w_j
\ldots w_{i-1} \cdot w_i A''$ in $s_{i-1}$.  The rest of the
conditions are exactly as in the previous case.

Thus, if $w = w_1 \ldots w_n$ is accepted by $\fsa$, then there is a
path from $s_0$ to $s_n$ labeled by $w_1 \ldots w_n$.  Hence, by the
claim just proved, $A \ra x \cdot yB$ is a dotted rule of $s_n$, and
$B \derives z$, where $yz = w_1 \ldots w_n = w$.  Because the $s_i$ in
the claim is $s_0$, and all the dotted rules of $s_i$ can have nothing
before the dot, and $x$ must be the empty string.  Therefore, the only
possible case is case 3.  Thus, $S
\derives A \ra yz = w$, and hence $w \in L(G)$.
The proof that the empty string is accepted by $\fsa$
only if it is in $L(G)$ is similar to the proof of the claim.
\end{proof}

\subsection{Decompositions}

If each $\const{def}(X)$ in the strongly-connected component
decomposition of $G$ is left-linear or right-linear, it is easy to see
that $G$ accepts a regular language, and that the overall
approximation derived by decomposition is exact. Since some
components may be left-linear and others right-linear, the overall
class we can approximate exactly goes beyond purely left-linear or
purely right-linear grammars.

\section{Implementation and Example}
\begin{table} 
\begin{center}
\begin{tabular}{|l|l|l|}
\hline
	Symbol	&	Category	& Features \\
\hline
	{\tt s}	&	sentence	& {\tt n} (number), {\tt p} (person) \\
	{\tt np}&	noun phrase	& {\tt n}, {\tt p}, {\tt c} (case)\\
	{\tt vp}	&	verb phrase	& {\tt n}, {\tt p},
{\tt t} (verb type)\\
	{\tt args}	&	verb arguments	& {\tt t}\\
	{\tt det}	&	determiner	& {\tt n}\\
	{\tt n}	&	noun		& n\\
	{\tt pron}	&	pronoun		& {\tt n}, {\tt p},
{\tt c}\\
	{\tt v}	& 	verb		& {\tt n}, {\tt p}, {\tt t}\\
\hline
\end{tabular}
\end{center}
\caption{Categories of Example Grammar}
\label{tab1}
\end{table}
\begin{table}
\begin{center}
\begin{tabular}{|l|p{2in}|}
\hline
	Feature		& Values \\
\hline
	{\tt n} (number)	& {\tt s} (singular), {\tt p} (plural) \\
	{\tt p} (person)	& {\tt 1} (first), {\tt 2} (second),
{\tt 3} (third) \\
	{\tt c} (case)	& {\tt s} (subject), {\tt o} (nonsubject) \\
	{\tt t} (verb type)	& {\tt i} (intransitive), {\tt t}
(transitive), {\tt d}
(ditransitive) \\
\hline
\end{tabular}
\end{center}
\caption{Features of Example Grammar}
\label{tab2}
\end{table}
The example in the appendix is an APSG for a small fragment
of English, written in the notation accepted by our grammar compiler.
The categories and features used in the grammar are described in
Tables \ref{tab1} and
\ref{tab2} (categories without features are omitted).
The example grammar accepts sentences such as 
\begin{verse}
\tt i give a cake to tom \\
tom sleeps \\
i eat every nice cake
\end{verse}
but rejects ill-formed inputs such as
\begin{verse}
\tt  i sleeps \\
i eats a cake \\
i give \\
tom eat
\end{verse}
\noindent It is easy to see that the each strongly-connected component
of the example is either left-linear or right linear, and therefore
our algorithm will produce an equivalent FSA.
\begin{figure}
\centerline{\psfig{figure=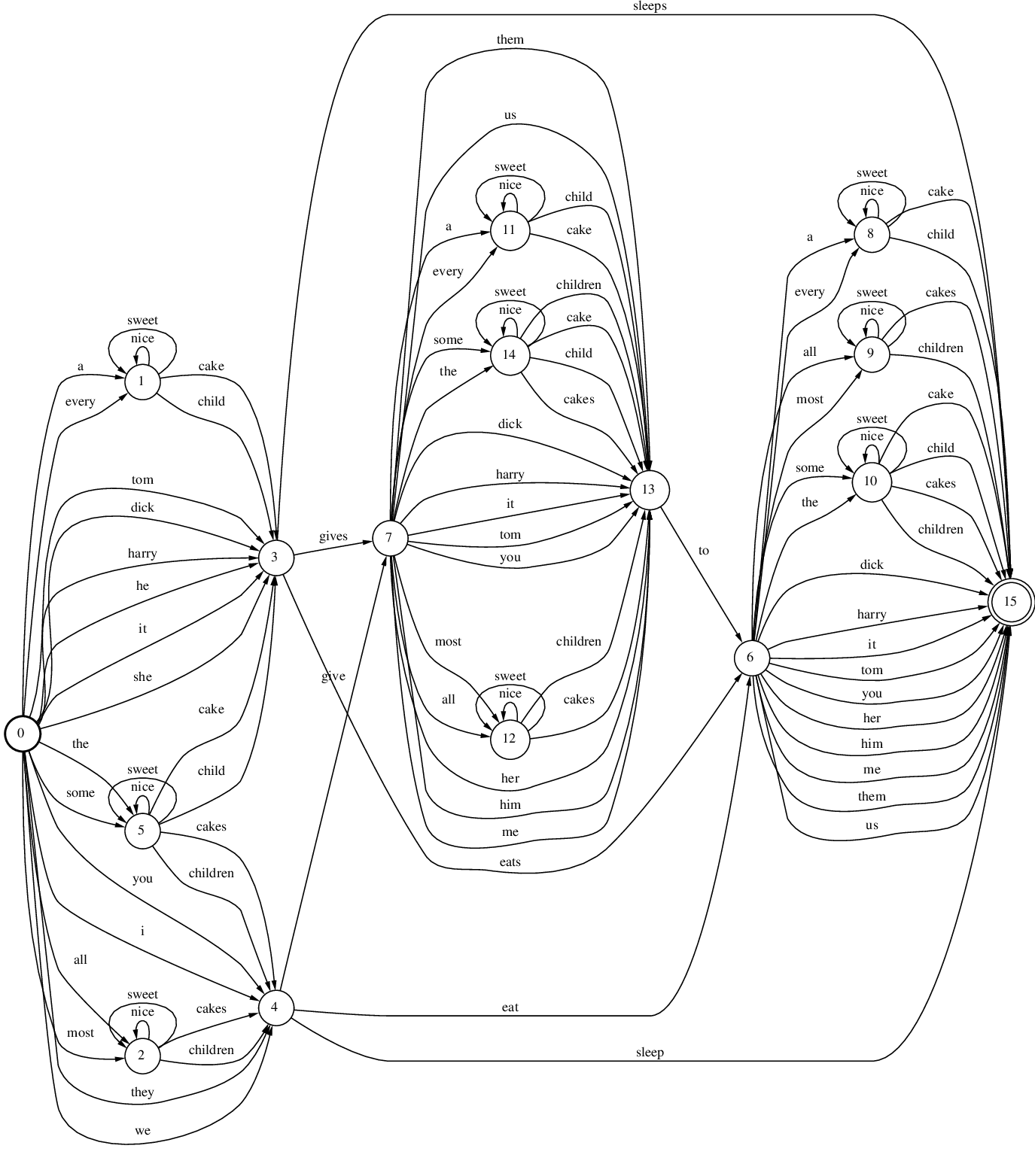,width=6in}}
\caption{Approximation for Example Grammar}
\label{big-dfa}
\end{figure}
{\samepage
Grammar compilation is organized as follows:
\begin{enumerate}
\item Instantiate input APSG to yield an equivalent CFG.
\item Decompose the CFG into strongly-connected components.
\item For each subgrammar $\const{def}(X)$ in the decomposition:
\begin{enumerate}
\item approximate $\const{def}(X)$ by $\const{aut}(X)$;
\item determinize and minimize $\const{aut}(X)$;
\end{enumerate}
\item Recombine the $\const{aut}(X)$ into a single FSA using the
partial order of grammar components.
\item Determinize and minimize the recombined FSA.
\end{enumerate}
}
\noindent For small examples such as the present one, steps 2, 3 and 4 can be
replaced by a single approximation step for the whole CFG.  In the
current implementation, instantiation of the APSG into an equivalent
CFG is written in Prolog, and the other compilation steps are written
in C, for space and time efficiency in dealing with potentially large
grammars and automata.

For the example grammar, the equivalent CFG has 78 nonterminals and
157 rules, the unfolded and flattened FSA 2615 states and 4096
transitions, and the determinized and minimized final DFA shown in
Figure \ref{big-dfa} has 16 states and 97 transitions. The runtime for
the whole process is 1.78 seconds on a Sun SparcStation 20.

Substantially larger grammars, with thousands of instantiated rules,
have been developed for a speech-to-speech translation project
\cite{Roe+al-92:VEST}. Compilation times vary widely, but very long
compilations appear to be caused by a combinatorial explosion in the
unfolding of right recursions that will be discussed further in the
next section.

\section{Informal Analysis}

In addition to the cases of left-linear and right-linear grammars and
decompositions into those cases discussed in Section \ref{formal}, our
algorithm is exact in a variety of interesting cases, including the
examples of Church and Patil
\cite{Church+Patil:ambiguity}, which illustrate how typical attachment
ambiguities arise as structural ambiguities on regular string sets.

The algorithm is also exact for some self-embedding grammars\footnote{A
grammar is self-embedding if and only if licenses the derivation
$X\derives\alpha X\beta$ for nonempty $\alpha$ and
$\beta$. A language is regular if and only if it can be described by some
non-self-embedding grammar.} of regular languages, such as
\[
S\ra a S \mid  S b \mid  c
\]
defining the regular language $a^{*}cb^{*}$. 

\begin{figure}
\centerline{\psfig{figure=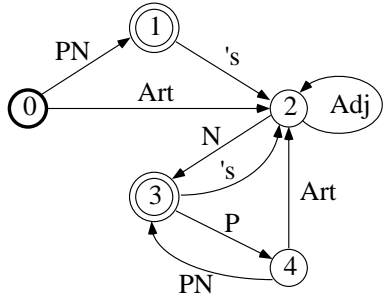,width=2in}}
\caption{Acceptor for Noun Phrases}
\label{fig4}
\end{figure}

A more interesting example is the following simplified grammar
for the structure of English noun phrases:
\[
\begin{array}{l}
\mbox{NP}\ra\mbox{Det}\;\mbox{Nom} \mid  \mbox{PN}\\
\mbox{Det}\ra\mbox{Art} \mid  \mbox{NP}\;\mbox{'s} \\
\mbox{Nom}\ra\mbox{N} \mid  \mbox{Nom}\;\mbox{PP} \mid 
\mbox{Adj}\;\mbox{Nom} \\
\mbox{PP}\ra\mbox{P}\;\mbox{NP}
\end{array}
\]
The symbols Art, Adj, N, PN and P correspond to the parts of speech
article, adjective, noun, proper noun and preposition, and the
nonterminals Det, NP,
Nom and PP to determiner phrases, noun phrases, nominal phrases and
prepositional phrases, respectively. From this grammar, the
algorithm derives the exact DFA in Figure \ref{fig4}.
This example is typical of the kinds of grammars with systematic
attachment ambiguities discussed by Church and Patil
\cite{Church+Patil:ambiguity}. A string of parts-of-speech such as
\[
\const{Art}\;\const{N}\;\const{P}\;\const{Art}\;\const{N}\;\const{P}\;
\const{Art}\;\const{N}
\]
is ambiguous according to the grammar (only some constituents shown
for simplicity):
\[
\begin{array}{l}
\const{Art}\; [_{\const{\scriptsize{}Nom}} \const{N}\;[_{\const{\scriptsize{}PP}}\const{P}
[_{\const{\scriptsize{}NP}}  \const{Art}\; [_{\const{\scriptsize{}Nom}}\;\const{N}\;
[_{\const{\scriptsize{}PP}}\const{P}
\;[_{\const{\scriptsize{}NP}}\const{Art}\;\const{N}]]]]]] \\
\const{Art}\; [_{\const{\scriptsize{}Nom}}  [_{\const{\scriptsize{}Nom}}
\const{N}\;[_{\const{\scriptsize{}PP}} \const{P}\;[_{\const{\scriptsize
NP}} \const{Art}\;\const{N}]]]\; [_{\const{\scriptsize{}PP}}\const{P}\;
[_{\const{\scriptsize{}NP}}\const{Art}\;\const{N}]]]
\end{array}
\]
However, if multiplicity of analyses are ignored, the string set accepted
by the grammar is regular and the approximation algorithm obtains the
correct DFA. However, we have no characterization of the class of CFGs
for which this kind of exact approximation is possible.

As an example of inexact approximation, consider the self-embedding CFG
\[
S \ra a S b \mid  \epsilon 
\]
for the nonregular language $a^n b^n, n\ge 0$. This grammar is mapped
by the algorithm into an FSA accepting $\epsilon \mid a^{+}b^{+}$. The
effect of the algorithm is thus to ``forget'' the pairing between
$a$'s and $b$'s mediated by the stack of the grammar's characteristic
recognizer. 

Our algorithm has very poor worst-case performance. First,
the expansion of an APSG into a CFG, not described here, can lead
to an exponential blow-up in the number of nonterminals and rules.
Second, the subset calculation implicit in the LR(0) construction can
make the number of states in the characteristic machine exponential on
the number of CF rules. Finally, unfolding can yield another
exponential blow-up in the number of states. 

However, in the practical
examples we have considered, the first and the last problems appear to
be the most serious.

The rule instantiation problem may be alleviated by avoiding full
instantiation of unification grammar rules with respect to ``don't
care'' features, that is, features that are not constrained by the
rule.

The unfolding problem is particularly serious in grammars with
subgrammars of the form
\begin{equation}S \ra X_1 S \mid \cdots \mid X_n S \mid
Y\qquad\mbox{.}\label{right-blowup}\end{equation} It is easy to see
that the number of unfolded states in the subgrammar is exponential in
$n$. This kind of situation often arises indirectly in the expansion
of an APSG when some features in the right-hand side of a rule are
unconstrained and thus lead to many different instantiated rules.
However, from the proof of Proposition \ref{right-linear} it follows
immediately that unfolding is unnecessary for right-linear grammars.
Therefore, if we use our grammar decomposition method first and test
individual components for right-linearity, unnecessary unfolding can
be avoided. Alternatively,
the problem can
be circumvented by left factoring (\ref{right-blowup}) as follows:
\[
\begin{array}{l}
S \ra Z S \mid  Y \\
Z \ra X_1 \mid  \cdots \mid  X_n
\end{array}
\]

\section{Related Work and Conclusions}

Our work can be seen as an algorithmic realization of suggestions of
Church and Patil \cite{Church:masters,Church+Patil:ambiguity} on
algebraic simplifications of CFGs of regular languages. Other work on
finite state approximations of phrase structure grammars has typically
relied on arbitrary depth cutoffs in rule application. While this may
be reasonable for psycholinguistic modeling of performance
restrictions on center embedding \cite{Pulman:memory}, it does not
seem appropriate for speech recognition where the approximating FSA is
intended to work as a filter and not reject inputs acceptable by the
given grammar. For instance, depth cutoffs in the method described by
Black \cite{Black:fsm} lead to approximating FSAs whose language is
neither a subset nor a superset of the language of the given
phrase-structure grammar. In contrast, our method will produce an
exact FSA for many interesting grammars generating regular languages,
such as those arising from systematic attachment ambiguities
\cite{Church+Patil:ambiguity}.  It is important to note, however, that
even when the result FSA accepts the same language, the original
grammar is still necessary because interpretation algorithms are
generally expressed in terms of phrase structures described by that
grammar, not in terms of the states of the FSA.

Several extensions of the present work may be worth investigating.

As is well known, speech recognition accuracy can often be improved by
taking into account the probabilities of different sentences. If such
probabilities are encoded as rule probabilities in the initial
grammar, we would need a method for transferring them to the
approximating FSA. Alternatively, transition probabilities for the
approximating FSA could be estimated directly from a training corpus,
either by simple counting in the case of a DFA or by an appropriate
version of the Baum-Welch procedure for general probabilistic FSAs
\cite{Rabiner:HMM-tutorial}.

Alternative pushdown acceptors and stack congruences may be
considered with different size-accuracy tradeoffs. Furthermore,
instead of expanding the APSG first into a CFG and only then
approximating, one might start with a pushdown acceptor for the APSG
class under consideration \cite{Lang:LPDA}, and approximate it
directly using a generalized notion of stack congruence that takes
into account the instantiation of stack items.  This approach might
well reduce the explosion in grammar size induced by the initial
conversion of APSGs to CFGs, and also make the method applicable to
APSGs with unbounded feature sets, such as general constraint-based
grammars.

We do not have any useful quantitative measure of approximation
quality. Formal-language theoretic notions such as the rational index
of a language \cite{Boasson+al:rational-index} capture a
notion of language complexity but it is not clear how it relates to
the intuition that an approximation is ``worse'' than another
if it strictly contains it. In a probabilistic
setting, a language can be identified with a probability density
function over strings. Then the Kullback-Leibler divergence
\cite{Cover+Thomas:info} between the approximation and the original
language might be a useful measure of approximation quality.

Finally, constructions based on finite-state transducers may lead to a
whole new class of approximations. For instance, CFGs may be
decomposed into the composition of a simple fixed CFG with given
approximation and a complex, varying finite-state transducer that
needs no approximation.

\section*{Acknowledgments}
We thank Mark Liberman for suggesting that we look into finite-state
approximations, Richard Sproat, David Roe and Pedro Moreno trying out
several prototypes supplying test grammars, and Mehryar Mohri, Edmund
Grimley-Evans and the editors of this volume for corrections and other
useful suggestions. This paper is a revised and extended version of
the 1991 ACL meeting paper with the same title
\cite{Pereira+Wright:acl}.

\bibliographystyle{plain}
\bibliography{new}

\begin{thebibliography}{10}

\bibitem{Aho+Hopcroft+Ullman:da}
Alfred~V. Aho, John~E. Hopcroft, and Jeffrey~D. Ullman.
\newblock {\em The Design and Analysis of Computer Algorithms}.
\newblock Addison-Wesley, Reading, Massachusetts, 1976.

\bibitem{Aho+Ullman:principles}
Alfred~V. Aho and Jeffrey~D. Ullman.
\newblock {\em Principles of Compiler Design}.
\newblock Addison-Wesley, Reading, Massachusetts, 1977.

\bibitem{Backhouse:syntax}
Roland~C. Backhouse.
\newblock {\em Syntax of Programming Languages---Theory and Practice}.
\newblock Series in Computer Science. Prentice-Hall, Englewood Cliffs, New
  Jersey, 1979.

\bibitem{Black:fsm}
Alan~W. Black.
\newblock Finite state machines from feature grammars.
\newblock In Masaru Tomita, editor, {\em International Workshop on Parsing
  Technologies}, pages 277--285, Pittsburgh, Pennsylvania, 1989. Carnegie
  Mellon University.

\bibitem{Boasson+al:rational-index}
Luc Boasson, Bruno Courcelle, and Maurice Nivat.
\newblock The rational index: a complexity measure for languages.
\newblock {\em {SIAM} Journal of Computing}, 10(2):284--296, 1981.

\bibitem{Carpenter:logic}
Bob Carpenter.
\newblock {\em The Logic of Typed Feature Structures}.
\newblock Number~32 in Cambridge Tracts in Theoretical Computer Science.
  Cambridge University Press, Cambridge, England, 1992.

\bibitem{Church:masters}
Kenneth~W. Church.
\newblock On memory limitations in natural language processing.
\newblock Master's thesis, M.I.T., 1980.
\newblock Published as Report MIT/LCS/TR-245.

\bibitem{Church+Patil:ambiguity}
Kenneth~W. Church and Ramesh Patil.
\newblock Coping with syntactic ambiguity or how to put the block in the box on
  the table.
\newblock {\em Computational Linguistics}, 8(3--4):139--149, 1982.

\bibitem{Cover+Thomas:info}
Thomas~M. Cover and Joy~A. Thomas.
\newblock {\em Elements of Information Theory}.
\newblock Wiley-Interscience, New York, New York, 1991.

\bibitem{Lang:LPDA}
Bernard Lang.
\newblock Complete evaluation of {Horn} clauses: an automata theoretic
  approach.
\newblock Rapport de Recherche 913, {INRIA}, Rocquencourt, France, November
  1988.

\bibitem{Pereira+Wright:acl}
Fernando C.~N. Pereira and Rebecca~N. Wright.
\newblock Finite-state approximation of phrase-structure grammars.
\newblock In {\em 29th Annual Meeting of the Association for Computational
  Linguistics}, pages 246--255, Berkeley, California, 1991. University of
  California at Berkeley, Association for Computational Linguistics,
  Morristown, New Jersey.

\bibitem{Pulman:memory}
Steven~G. Pulman.
\newblock Grammars, parsers, and memory limitations.
\newblock {\em Language and Cognitive Processes}, 1(3):197--225, 1986.

\bibitem{Rabiner:HMM-tutorial}
Lawrence~R. Rabiner.
\newblock A tutorial on hidden markov models and selected applications in
  speech recognition.
\newblock {\em Proceedings of the {IEEE}}, 77(2):257--286, 1989.

\bibitem{Roe+al-92:VEST}
David~B. Roe, Pedro~J. Moreno, Richard~W. Sproat, Fernando C.~N. Pereira,
  Michael~D. Riley, and Alejandro Macarr\'{o}n.
\newblock A spoken language translator for restricted-domain context-free
  languages.
\newblock {\em Speech Communication}, 11:311--319, 1992.

\bibitem{Shieber:intro}
Stuart~M. Shieber.
\newblock {\em An Introduction to Unification-Based Approaches to Grammar}.
\newblock Number~4 in CSLI Lecture Notes. Center for the Study of Language and
  Information, Stanford, California, 1985.
\newblock Distributed by Chicago University Press.

\bibitem{Shieber:restriction}
Stuart~M. Shieber.
\newblock Using restriction to extend parsing algorithms for
  complex-feature-based formalisms.
\newblock In {\em 23rd Annual Meeting of the Association for Computational
  Linguistics}, pages 145--152, Chicago, Illinois, 1985. Association for
  Computational Linguistics, Morristown, New Jersey.

\bibitem{Ullian:partial-cfls}
Joseph~S. Ullian.
\newblock Partial algorithm problems for context free languages.
\newblock {\em Information and Control}, 11:90--101, 1967.

\end{thebibliography}

\section*{Appendix---APSG Formalism and Example}
Nonterminal symbols (syntactic categories) may
have features that specify variants of the category (eg. singular or
plural noun phrases, intransitive or transitive verbs). A category
{\em cat}
with feature constraints is written
\[\mbox{\em cat}\mbox{\tt \#[}c_1,\ldots,c_m\mbox{\tt ]} \mbox{\tt .}\]

Feature constraints for feature $f$ have one of the forms
\begin{eqnarray}
f & \mbox{\tt =} &  v \label{eqn-1}\\
f & \mbox{\tt =} &  c \label{eqn-2}\\
f & \mbox{\tt =} &  \mbox{\tt (} c_1, \ldots , c_n \mbox{\tt )}\label{eqn-3}
\end{eqnarray}
where $v$ is a variable name (which must be capitalized) and $c,
c_1,\ldots,c_n$ are feature values. 

All occurrences of a variable $v$ in a rule stand for the same
unspecified value. A constraint with form (\ref{eqn-1}) specifies a
feature as having that value. A constraint of form (\ref{eqn-2})
specifies an actual value for a feature, and a constraint of form
(\ref{eqn-3}) specifies that a feature may have any value from the
specified set of values.  The symbol ``!'' appearing as the value of a
feature in the right-hand side of a rule indicates that that feature
must have the same value as the feature of the same name of the
category in the left-hand side of the rule. This notation, as well as
variables, can be used to enforce feature agreement between categories
in a rule, for instance, number agreement between subject and verb.

It is convenient to declare the features and possible values of
categories with category declarations appearing before the grammar
rules. Category declarations have the form
\[
\mbox{\tt cat}\;\;\mbox{\em cat}\mbox{\tt \#[}
\begin{array}[t]{rcl}
f_1 & \mbox{\tt =} & \mbox{\tt (} v_{11},\ldots, v_{1k_{1}}\mbox{\tt
)}, \\
\multicolumn{3}{c}{\ldots ,} \\ 
f_m & \mbox{\tt =} & \mbox{\tt (} v_{m1}, \ldots, v_{mk_m}\mbox{\tt )}\quad\mbox{\tt ].}
\end{array}
\]
giving all the possible values of all the features for the category.

The declaration
\[\mbox{\tt start}\;\;\mbox{\em cat}\mbox{\tt .}\]
declares {\em cat} as the start symbol of the grammar.

In the grammar rules, the symbol
``{\tt `}'' prefixes terminal symbols, commas are used for sequencing
and ``{\tt |}'' for alternation.

\begin{verbatim}

start s.

cat s#[n=(s,p),p=(1,2,3)].
cat np#[n=(s,p),p=(1,2,3),c=(s,o)].
cat vp#[n=(s,p),p=(1,2,3),type=(i,t,d)].
cat args#[type=(i,t,d)].

cat det#[n=(s,p)].
cat n#[n=(s,p)].
cat pron#[n=(s,p),p=(1,2,3),c=(s,o)].
cat v#[n=(s,p),p=(1,2,3),type=(i,t,d)].

s => np#[n=!,p=!,c=s], vp#[n=!,p=!].

np#[p=3] => det#[n=!], adjs, n#[n=!].
np#[n=s,p=3] => pn.
np => pron#[n=!, p=!, c=!].

pron#[n=s,p=1,c=s] => `i.
pron#[p=2] => `you.
pron#[n=s,p=3,c=s] => `he | `she.
pron#[n=s,p=3] => `it.
pron#[n=p,p=1,c=s] => `we.
pron#[n=p,p=3,c=s] => `they.
pron#[n=s,p=1,c=o] => `me.
pron#[n=s,p=3,c=o] => `him | `her.
pron#[n=p,p=1,c=o] => `us.
pron#[n=p,p=3,c=o] => `them.

vp => v#[n=!,p=!,type=!], args#[type=!].

adjs => [].
adjs => adj, adjs.

args#[type=i] => [].
args#[type=t] => np#[c=o].
args#[type=d] => np#[c=o], `to, np#[c=o].

pn => `tom | `dick | `harry.

det => `some| `the.
det#[n=s] => `every | `a.
det#[n=p] => `all | `most.

n#[n=s] => `child | `cake.
n#[n=p] => `children | `cakes.

adj => `nice | `sweet.

v#[n=s,p=3,type=i] => `sleeps.
v#[n=p,type=i] => `sleep.
v#[n=s,p=(1,2),type=i] => `sleep.
  
v#[n=s,p=3,type=t] => `eats.
v#[n=p,type=t] => `eat.
v#[n=s,p=(1,2),type=t] => `eat.

v#[n=s,p=3,type=d] => `gives.
v#[n=p,type=d] => `give.
v#[n=s,p=(1,2),type=d] => `give.
\end{verbatim}

\end{document}